\documentclass{article}

\usepackage{graphicx} 
\usepackage{times}    
\usepackage{url}      
\usepackage{multirow}
\usepackage{listings}
\usepackage{algorithmic}
\usepackage{algorithm2e}
\usepackage{flushend}

\begin{document}

\title{SMURFF: a High-Performance Framework for Matrix Factorization}

\author{
Tom Vander Aa, Imen Chakroun and Thomas J. Ashby\\
ExaScience Life Lab at imec, Leuven, Belgium\\
\\
Jaak Simm, Adam Arany and Yves Moreau\\
ESAT-STADIUS, KU Leuven, Leuven, Belgium\\
\\
Thanh Le Van, Jos\'e Felipe Golib Dzib, J\"org Wegner,\\Vladimir Chupakhin and Hugo Ceulemans\\
Janssen Pharmaceutica, Beerse, Belgium\\
\\
Roel Wuyts and Wilfried Verachtert\\
ExaScience Life Lab at imec, Leuven, Belgium\\
\\
\emph{tom.vanderaa@imec.be}\\
}

\maketitle

\begin{abstract}
    Bayesian Matrix Factorization (BMF) is a powerful technique for recommender
    systems because it produces good results and is relatively robust against
    overfitting. Yet BMF is more computationally intensive and thus more
    challenging to implement for large datasets. In this work we present SMURFF a
    high-performance feature-rich framework to compose and construct different
    Bayesian matrix-factorization methods.  The framework has been successfully used in
    to do large scale runs of compound-activity prediction.  SMURFF is
    available as open-source and can be used both on a supercomputer and on a
    desktop or laptop machine. Documentation and several examples are provided
    as Jupyter notebooks using SMURFF's high-level Python API.
\end{abstract}

\section{Matrix Factorization}
\label{sec:intro}

Recommender Systems (RS) have become very common in recent years and are useful
in various real-life applications.  The most popular ones are probably
suggestions for movies on Netflix and books for
Amazon~\cite{Gomez-Uribe:2015:NRS:2869770.2843948}. However, they can also be
used in more unlikely area such drug discovery where a key problem is the
identification of candidate molecules that affect proteins associated with
diseases~\cite{Bredel2004}. 

\begin{figure}
    \centering
    \includegraphics[width=\columnwidth]{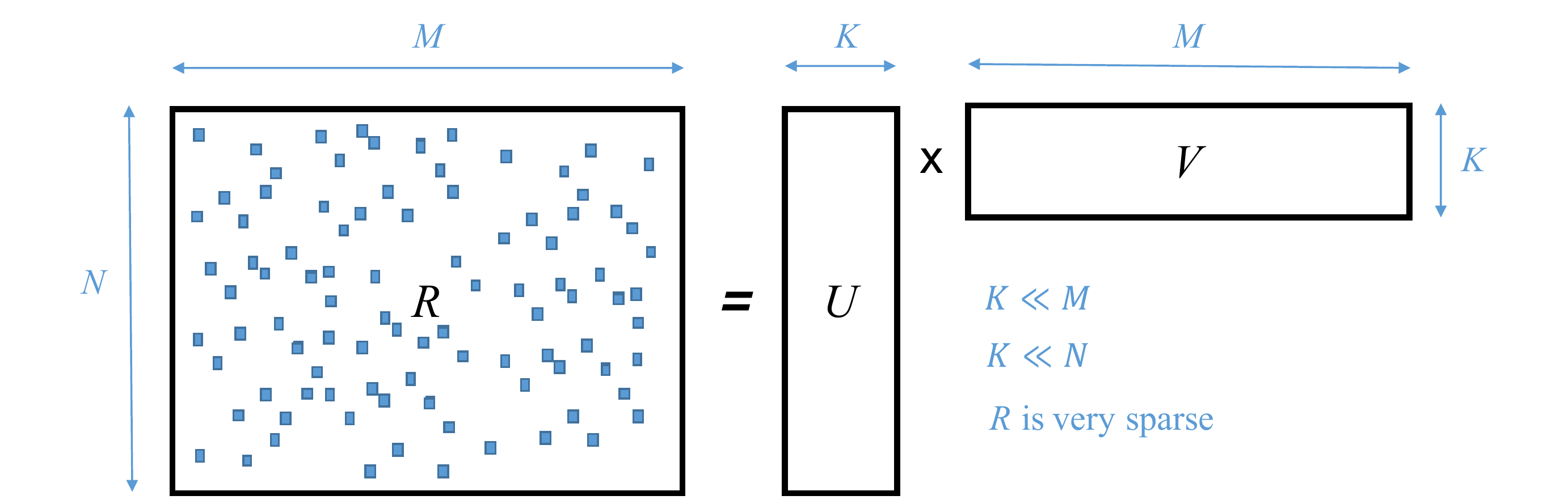}
    \caption{Low-rank Matrix Factorization}
    \label{fig:mf}
\end{figure}

In RS one has to analyze large and sparse matrices, for example those
containing the known movie or book ratings. Matrix Factorization (MF) is a
technique that has been successfully used here. As sketched in
Figure~\ref{fig:mf}, the idea of this method is to approximate the rating
matrix $R$ as a product of two low-rank matrices $U$ and $V$.  Predictions can
be made from the approximation $U \times V$ which is dense. 

Bayesian Matrix Factorization (BMF~\cite{BPMF}), using Gibbs Sampling is one of
the more popular approaches for matrix factorization. Thanks to the Bayesian
approach, BMF has been proven to be more robust to
data-overfitting~\cite{BPMF}. Gibbs sampling~\cite{gibbs} makes the problem
feasible to compute. Yet, BMF is still computationally intensive and thus more
challenging to implement for large datasets.

In this work we present SMURFF~\cite{smurff_github}, a high-performance
feature-rich framework to compose and construct different Bayesian
matrix factorization methods.  Using the SMURFF framework one
can easily vary: \emph{i)} the type of matrix to be factored (dense or sparse);
\emph{ii)} the prior-distribution that you assume the model to fit to
(multivariate-normal, spike-and-slab, and others); or \emph{iii)} the noise
model (Gaussian noise, Probit noise or others).  The framework also allows to
combine different matrices together and thus incorporate more and different
types of information into the model.  

In this paper we describe the SMURFF framework developed.
Section~\ref{sec:features} lists the different features supported. In
Section~\ref{sec:impl} we describe some of the implementation details and how
those impact compute performance. We compare SMURFF to other matrix
factorization implementations in Section~\ref{sec:results}, and look at compute
performance in Section~\ref{sec:perf} and Section~\ref{sec:conda}. Finally,
Section~\ref{sec:concl} presents conclusions and future work.

\section{Framework}
\label{sec:features}

Figure~\ref{fig:mf_framework} shows a matrix-factorization system as can be
composed with SMURFF. The basic matrix-components are:

\begin{figure}
    \centering
    \includegraphics[width=\columnwidth]{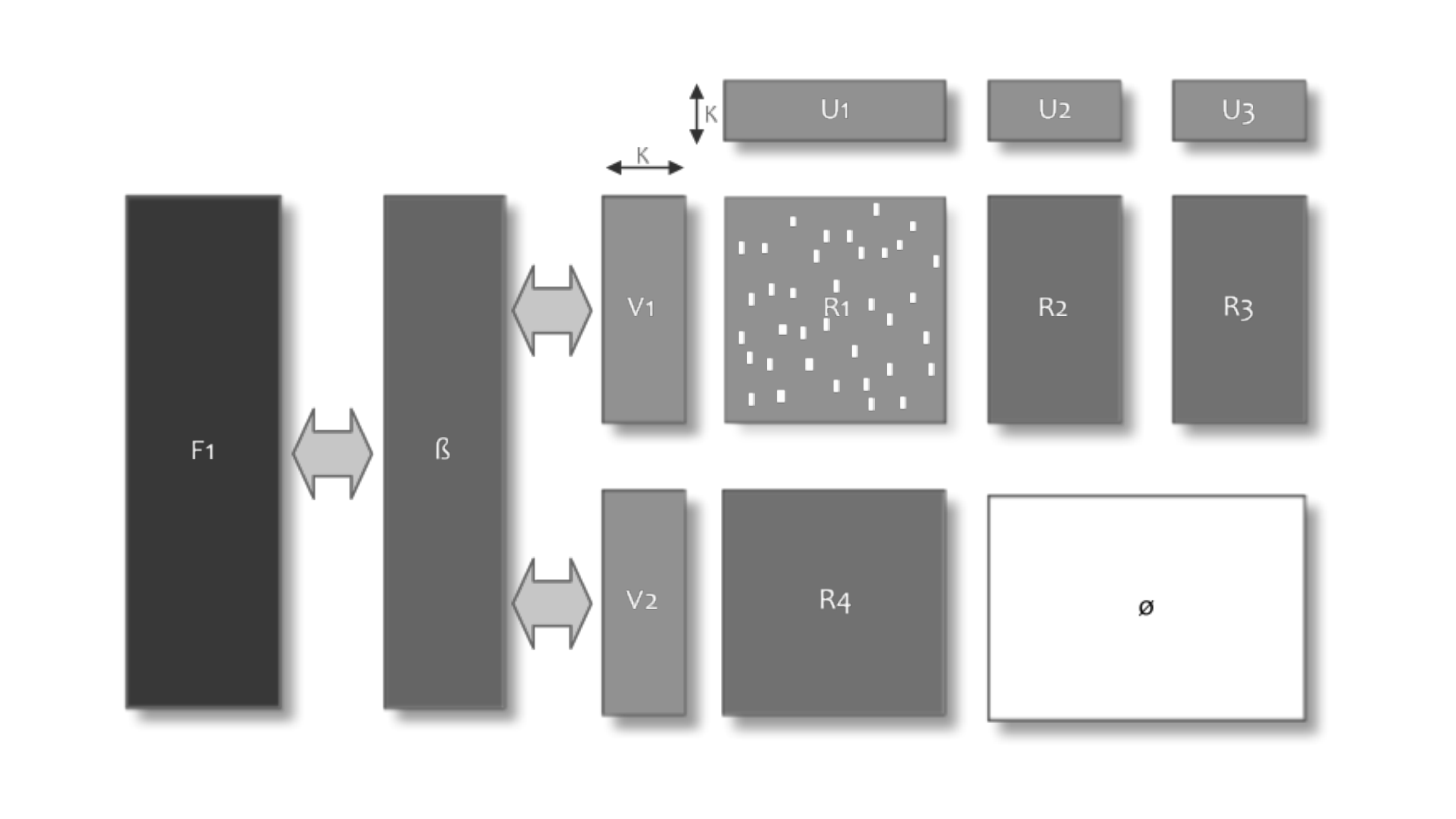}
    \caption{SMURFF Matrix Factorization Framework}
    \label{fig:mf_framework}
\end{figure}

\begin{itemize}
    \item The matrix to be factored $R$, which can be composed of different blocks
        ($R1$, $R2$, ...). Each of the blocks can be sparse with many values unknown, or
        fully known. If $Rx$ is fully known the data can still be sparse (many values 
        zero) or dense (all non-zeros).

    \item The model matrices, i.e. the factors $U$ and $V$. The rows in $U$ and 
        columns in $V$ is the so-called latent dimension of size $K$. Each user ($U$)
        each movie ($V$) is represented in by a latent vector of size $K$. Each element
        of such a vector is called a component.
        
    \item A matrix with side information ($F$) for the rows and/or columns of
        $R$ is taken into account by means of constructing a link-matrix
        ($\beta$). More information is in the paragraph below.
\end{itemize}

In the following paragraphs we describe the basic components of the matrix
factorization framework and how they are used or constructed.  The columns in
Table~\ref{tab:mf_algos} represent the different \emph{choices} SMURFF supports
in terms of input matrices, prior distributions, noise models and types of side
informations. We explain what those choices mean in the following paragraphs.

\begin{table*}[ht]
  \centering
  \footnotesize
  \caption{Possible MF Algorithms}
  \label{tab:mf_algos}
  \begin{tabular}{l|l|l|l|l}
      & \textbf{Input Matrices} & \textbf{Prior Distribution} & \textbf{Noise Model} & \textbf{Side Information} \\
      \hline
    \multirow{3}{*}{Choices}  & Sparse with unknowns & Normal             & Fixed Gaussian    & Link Matrix \\
                              & Sparse fully known   & Spike-and-Slab (SnS)     & Adaptive Gaussian & -  \\
                              & Dense-Dense          &                    &                   & - \\
    \hline
    BMF  & Sparse with unknowns & Normal & Fixed & - \\
    Macau & Sparse with unknowns & Normal & Fixed/Adaptive & Link Matrix \\
    GFA   & Sparse and/or Dense & Normal + SnS & Fixed/Adaptive &  -\\
    \hline
  \end{tabular}
\end{table*}

\paragraph{Input Matrices}

As already explained in Figure~\ref{fig:mf_framework}, different
matrix types are supported.

\paragraph{Prior Distributions}

Samples from $U$ / $V$ are taken from either a multi-variate normal
distribution or from a spike-and-slab distribution~\cite{spikeandslab}. The
spike-and-slab prior is a Gaussian distribution with an extra constraint
that enforces some latent components to zero. This allows to find out common and
disjoint factors between the different data sources~\cite{gfa}.

\paragraph{Noise model}

Noise is injected in the model when sampling. This noise is taken from a
Gaussian distribution, with either a fixed precision or with an adaptive
precision calculated based on how precise the model matches the data.

\paragraph{Side Information}

Side information are properties for each row or column of the $R$ matrix. For
example, for compound-activity prediction this could mean the physical shape or
chemical properties of the compounds can be used to improve the quality of the
factorization.

The link matrix $\beta$ to include the side information stored in $F$ in the
model, is constructed using the Macau algorithm described in \cite{simm:macau}. 

\paragraph{Possible Algorithms}

The aforementioned options on the type of input matrices, the prior
distribution, the noise model and inclusion of side-information through a link
matrix can be combined in different ways, resulting in different MF algorithms.
The bottom part of Table~\ref{tab:mf_algos} shows the choices to take to
implement Macau~\cite{simm:macau}, BMF~\cite{BPMF} and Group Factor Analysis
(GFA)~\cite{gfa}. Of course, other combinations are possible.

\section{High-Performance Implementation}
\label{sec:impl}

The most simple implementation (a single matrix, without side-information) can
be expressed as the pseudo code shown in Algorithm~\ref{algo:smurff_pseudo}. Most
time is spent in the loops updating $U$ and $V$, where each iteration consist
of some relatively basic matrix and vector operations on $K \times K$ matrices,
and one computationally more expensive $K \times K$ matrix inversion. 

\begin{algorithm}
  \LinesNumbered
  \For{sampling iterations}
  {
      sample hyper-parameters movies based on V

      \For{all movies $m$ of $M$}
      {
          update movie model $m$ based on
          ratings ($R$) for this movie and 
          model of users that rated this movie,
          plus randomly sampled noise
      }

      sample hyper-parameters users based on U

      \For{all users $u$ of $U$}
      {
          update user $u$ based on
          ratings ($R$) for this user and 
          model of movies this user rated,
          plus randomly sampled noise
      }

      \For{all test points}
      {
          predict rating and compute RMSE
      }

  }
  \caption{SMURFF Pseudo Code} 
  \label{algo:smurff_pseudo}
\end{algorithm}%

These matrix and vector operations are very well supported in
Eigen~\cite{Eigen} a high-performance modern C++11 linear algebra library.
Sampling from the basic distributions is available in the C++ standard template
library (STL), or can be trivially implemented on top. As a result the
Eigen-based C++ version of Algorithm~\ref{algo:smurff_pseudo} is a mere 35 lines
of C++ code. This implementation only supports BMF with a single sparse matrix and
has been the starting point of the SMURFF framework.

Although the structure of the fully-featured SMURFF framework is similar to
Algorithm~\ref{algo:smurff_pseudo} it consists of more than 25 000 lines of
code.

A Python API has been built on top of this C++ low-level library allowing
users of SMURFF to combine it with other Python packages for machine learning
and matrices such as \emph{numpy}, \emph{scipy} or
\emph{scikit-learn}~\cite{scikit-learn}. Several elaborate examples and a
tutorial are available using Jupyter notebooks~\cite{Kluyver2016JupyterN}.

\paragraph{Multi-core parallelisation}

Since the number of users and movies is very large and since all can be
computed in parallel, the \emph{for-all-users} and \emph{for-all-movies} loops
in Algorithm~\ref{algo:smurff_pseudo} become parallel-for loops. Inside a
single user/movie the amount of work depends on the number of non-zero entries
in $R$. If a certain user/movie has many non-zero entries, we use a parallel
algorithm, effectively splitting them up in more smaller tasks that can utilize
multiple cores on the system.

In the current implementation the parallelism is exploited using
OpenMP~\cite{openmp} parallel for loops, for the users/movie loops and OpenMP
tasks for the parallelization inside users/movies.

\section{Use Cases}
\label{sec:results}

In this section we look at the different matrix factorization methods that can 
be implemented using SMURFF, and compare the SMURFF and original implementations
with respect to compute performance and machine learning result.

\paragraph{Dataset} 
As a benchmark we use the ChEMBl~\cite{ChEMBL}, a dataset from the drug
discovery research field. ChEMBL contains descriptions for biological activities
involving over a million chemical entities, extracted primarily from scientific
literature. Several versions exist since the dataset is updated on a fairly
frequent basis. In this work, we use a dataset extracted from ChEMBL containing IC50
levels, which is a measure of the effectiveness of a substance in inhibiting a
specific biological or biochemical function, and ECFP chemical descriptors for
the compounds. 

\paragraph{BMF}

We compare the performance of the proposed multi-core BMF with several other
popular machine-learning packages that implement the BMF algorithm: We verified
that the predictive performance of the model, from all implementations is the
same.

\begin{figure}[h]
	\centering
	\includegraphics[width=\columnwidth]{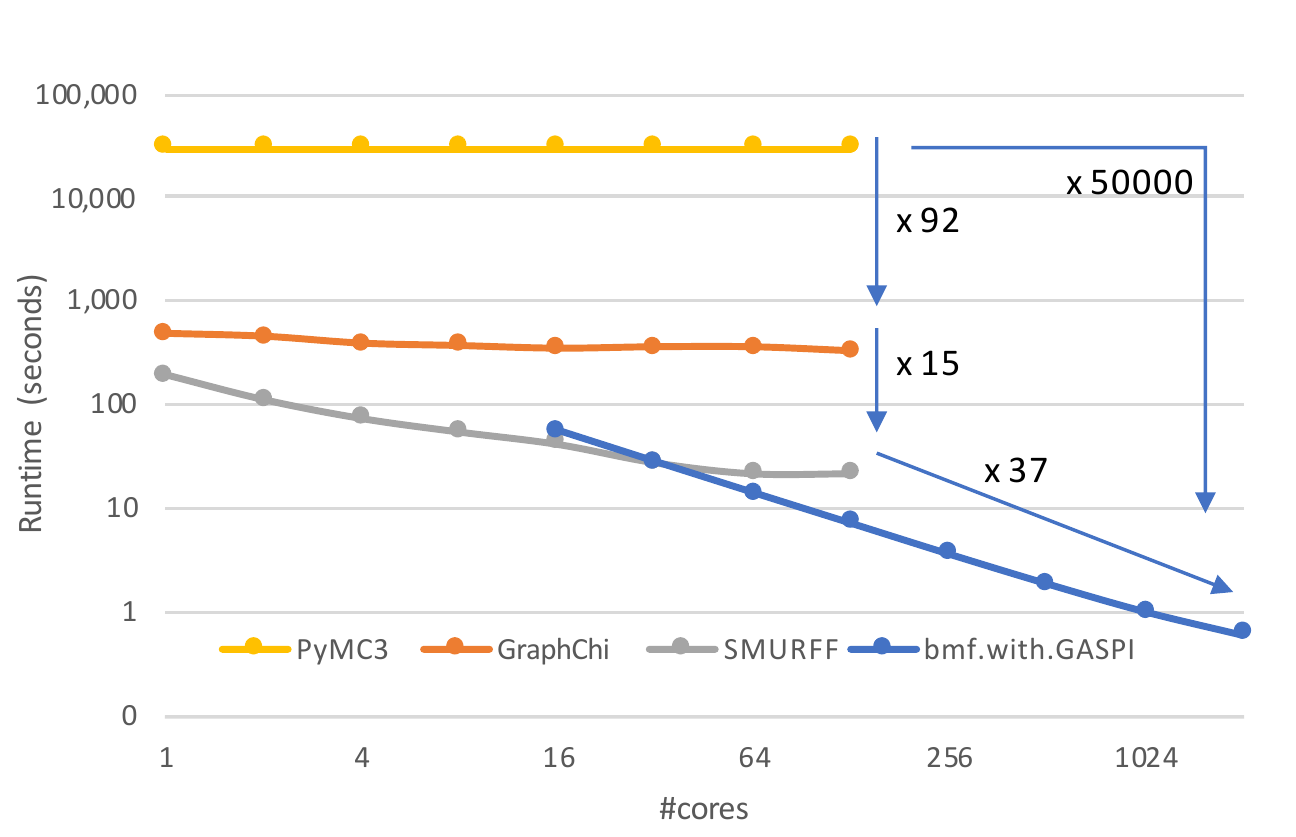}	
	\caption{Runtime of different BMF implementations.}
	\label{fig:pkgs}
\end{figure}

\begin{enumerate}
    \item \textbf{PyMC3}: PyMC3 is a Python package for Bayesian statistical
        modeling and probabilistic machine learning which focuses on advanced
        Markov chain Monte Carlo and variational fitting algorithms.
        PyMC3 relies on Theano for automatic
        differentiation and also for computation optimization and dynamic C
        compilation.~\cite{pymc3}

    \item \textbf{GraphChi}: Turi is a graph-based, high performance,
        distributed computation framework written in C++. In this graph we show
        the performance of GraphChi, a scalable high-performance multi-core
        implementation of Turi~\cite{graphchi}.

    \item \textbf{SMURFF}: this implementation.

    \item \textbf{BMF with GASPI}: A BMF-only multi-node
        implementation~\cite{bpmf_iccs17}. This code has been heavily optimized
        using the GASPI library~\cite{grunewald:gaspi} to run very efficient on
        multiple nodes. 

\end{enumerate}

Figure~\ref{fig:pkgs} shows the runtime of BMF~\cite{BPMF} for the four
implementations. The X-axis shows number of cores used. We have tested 
the PyMC3, GraphChi and SMURFF implementations on a single node with 36 cores, since
these implementations do not support multiple nodes, and the "BMF with
GASPI"-implementation on a system with 128 nodes and 2048 cores in total, since
this version has been optimized for a multi-node supercomputer.

The results show that SMURFF is $15\times$ faster compared to GraphChi, and even $1400\times$
compared to PyMC3. This is because PyMC3 and GraphChi are very versatile frameworks,
easy to use and with many more possibilities than SMURFF. PyMC3, for example,
has a high-level Python interface where one can choose many different samplers, 
from many different distributions. One can even implement custom samplers and
distributions in Python. SMURFF on the other hand only supports Gibbs sampling,
from Normal distributions.

As expected, the BMF-with-GASPI version scales very well, up to 1000 cores,
as we already showed in \cite{bpmf_iccs17}.

\paragraph{GFA}

We have explored the use and correctness of Group Factor Analysis (GFA) with
SMURFF by reproducing the results from \emph{Simulated study} in
\cite{DBLP:journals/corr/BunteLSK15}.


We have verified that SMURFF and the original R implementation of GFA produce
the same model, while the SMURFF C++ version is about $100\times$ faster. Using
the SMURFF implementation, we were able to reduce the runtime of GFA on a
realistic industrial dataset from 3 months for the R version \cite{gfa}, to 15
hours for the C++ implementation.  This speedup is expected, in general because
R is an interpreted language, but in this case especially since R is even
slower on sparse matrices and since the code contains many explicit
for-loops~\cite{for_loops}.


\paragraph{Macau}

The benefit of the Macau~\cite{simm:macau} algorithm is its ability to
incorporate side information for the rows and/or columns of the $R$ matrix.
SMURFF has been developed in collaboration with the authors of the Macau
implementation, and has the same code base. 

SMURFF with Macau has been used to predict compound-on-protein activity on a
matrix with more than one million compounds (rows) and several thousand
proteins (columns)~\cite{excape}. Chemical fingerprints were used for the
compounds, in both dense and sparse formats. This has led to important new
insights and potential new compounds to be used in drug discovery.

\section{Compute Performance}
\label{sec:perf}

In this section we look at the hardware performance of SMURFF on different
processor architectures and accelerators.  Since some hardware platforms are
better are handling sparse matrices, and some are better at dealing with dense
data, we also look at data types in this section.

\begin{figure}
	\centering
	\includegraphics[width=\columnwidth]{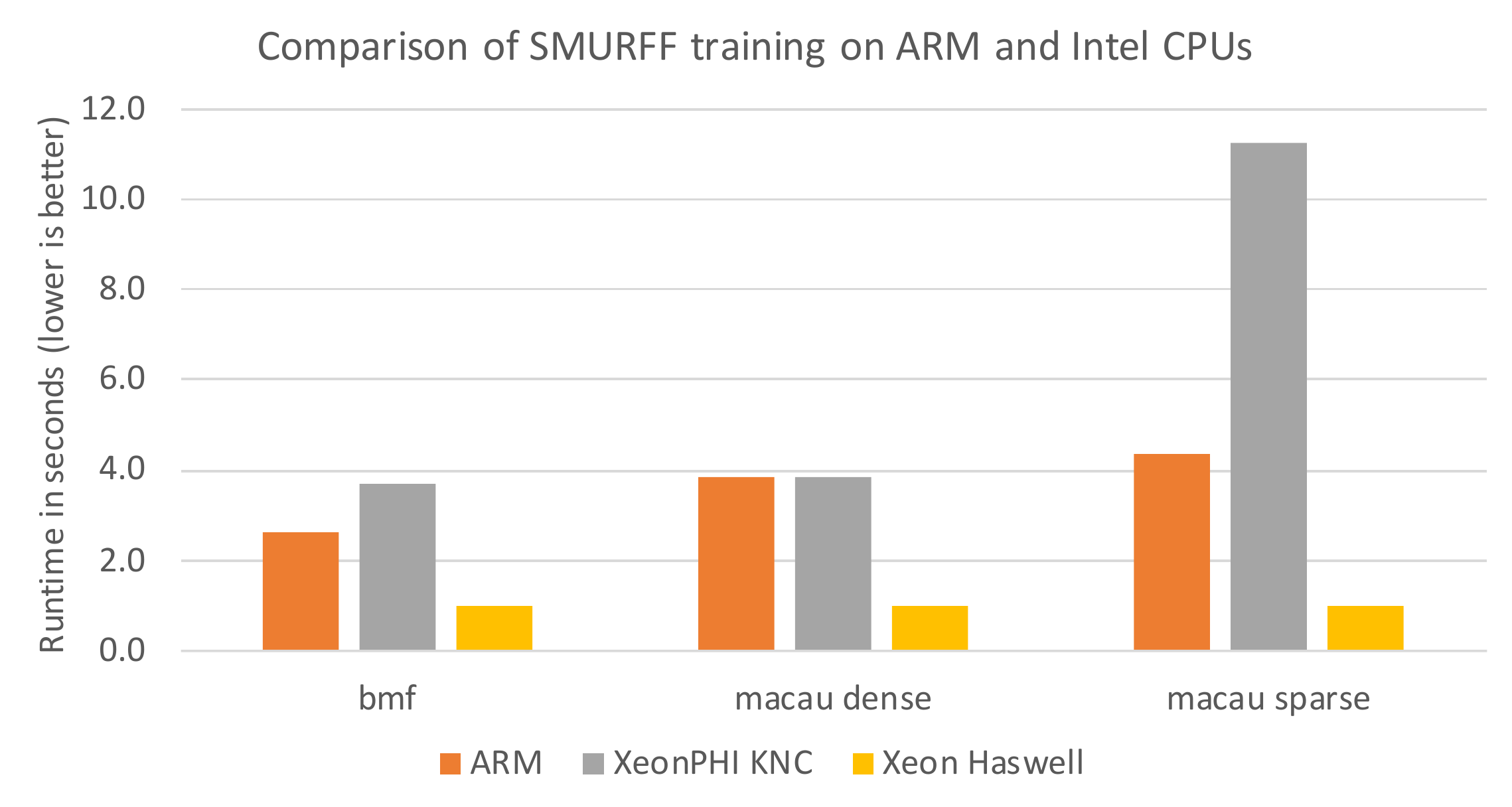}	
	\caption{Performance of BMF, Macau on dense input data 
        and Macau on sparse input data on three different hardware platforms.}
	\label{fig:hw}
\end{figure}

We compare three different hardware platforms:

\begin{enumerate}

        
    \item \textbf{Xeon}: a Haswell Intel Xeon processor running at 2.3 GHz with 36
        cores in 2 sockets ($2\times 18$ cores). We use up to 36 threads with Hyper Threading.


    \item \textbf{Xeon Phi}: an Intel KNC Xeon Phi processor running at 1.2 GHz wit
        61 cores. We use up to 244 threads with Hyper Threading.


    \item \textbf{ARM}: a ThunderX ARM 64bit processor with 96 cores.
\end{enumerate}

The results in Figure~\ref{fig:hw} show that the regular Xeon processor always
results in the best performance, and the \textbf{Xeon Phi} is always the worst,
the latter being between 4x and 10x slower. The two main reasons for the Xeon
Phi being slower are on one hand its lower clock frequency 1.2Ghz, compared to
3Ghz for the Xeon, and its inferior cache structure. It is known that the cache
coherency protocol on the regular Xeon is far better than the ring structure
that is used for L2 cache coherency on the Xeon
Phi~\cite{Fang:2014:TIX:2568088.2576799}.

The ARM processor performs in between the two Xeon processors, being on average
$3\times$ slower than the regular Xeon, sometimes closer to the Xeon Phi, sometimes
closer to the Xeon. This is explained by the difference in vector instruction
size. The Xeon uses 512bit AVX2 instructions, while the arm 128bit NEON
instructions.

The gap between the different hardware platforms is largest for sparse input
data, where the Xeon Haswell processor excels and the other two struggle. This
is mainly due to the superior memory architecture on the Xeon. The larger L3
cache of 40MB on the Xeon helps to keep the large sparse matrices, compared to
the smaller cache on the ARM (16MB) and the already mentioned crippled ring
interconnect between Xeon Phi cores.

\section{Binary Packaging using Conda}
\label{sec:conda}

\begin{figure}
	\centering
	\includegraphics[width=\columnwidth]{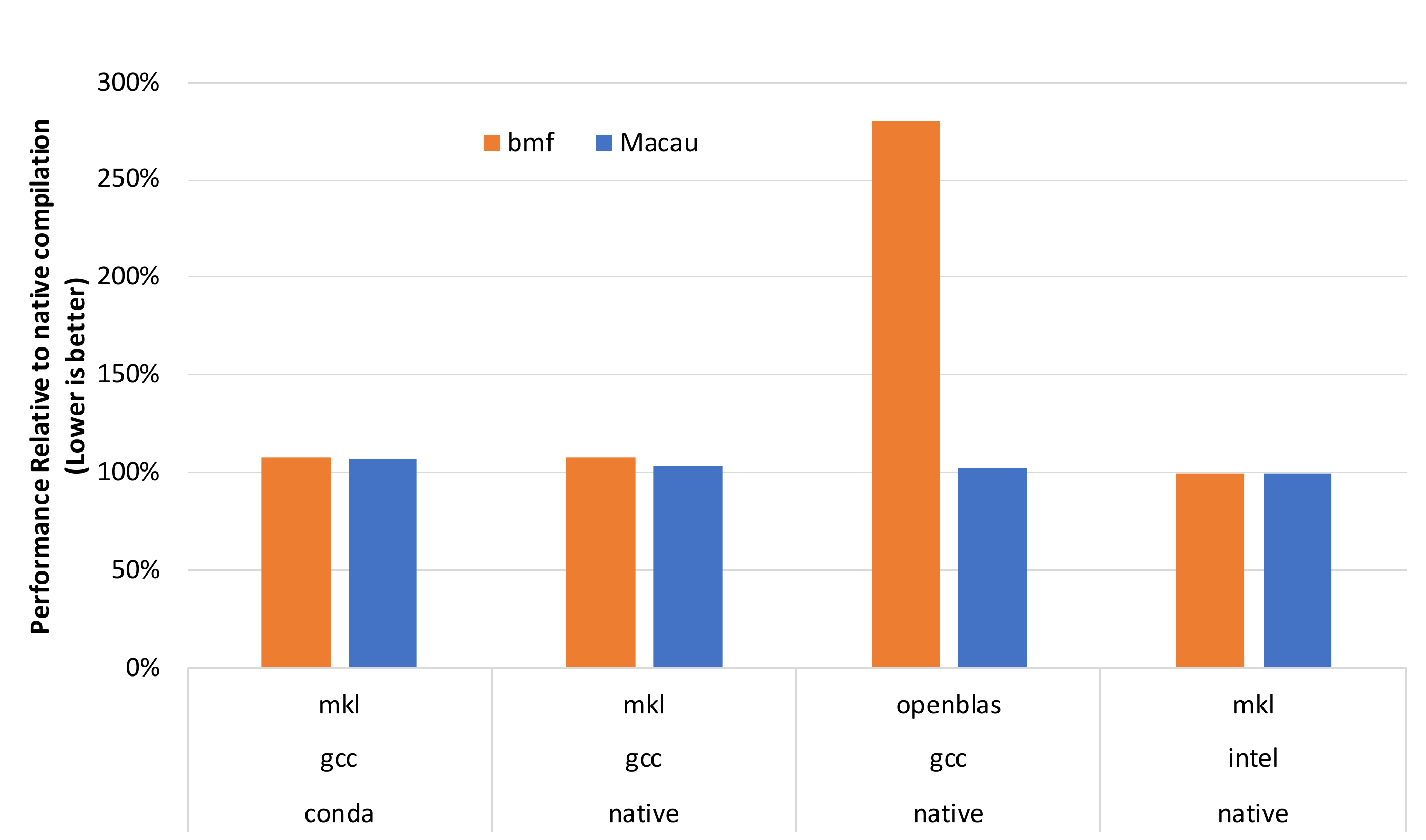}	
	\caption{Performance of compilation with Conda as opposed to native compilation.}
	\label{fig:conda}
\end{figure}

SMURFF can be installed with one command using Conda~\cite{smurff_conda}:
\texttt{conda install -c vanderaa smurff}

The main benefits of using Conda are:

\begin{itemize}
    \item Binary packages, no need to compile SMURFF yourself;
    \item Performance almost as good as compiling SMURFF yourself;
    \item A large ecosystem of packages, not only python;
    \item Works on Windows, Linux and Mac OS, with Python 2 and Python 3;
\end{itemize}

Figure~\ref{fig:conda} shows that it is indeed true that the binary Conda
package is compatible with many processor generations, without much loss in
performance. The figure shows runtime of four different combinations of
compilers (GCC or Intel compiler), algebra libraries (Intel Math Kernel
Library~\cite{mkl} or OpenBLAS) and target platform (either compilation for the
\emph{native} processor, or compilation for a generic hardware target with
\emph{Conda}.

Because the Intel MKL library dynamically adapts to the runtime hardware platform
it performs as good in \emph{native} as in \emph{Conda}-mode, and much better than
OpenBLAS, especially for the BMF algorithm. Using the Intel or GCC compiler 
does not make a difference, since both compilers are state-of-the-art and since
most of the time is spent in MKL.

\section{Conclusions and Future Work}
\label{sec:concl}

This paper described SMURFF a multi-core high-performance framework that
supports a wide range of Bayesian matrix-factorization algorithms, such as
Macau~\cite{simm:macau}, BMF~\cite{BPMF} and GFA~\cite{gfa}.  SMURFF is
available as open-source and can be used both on a supercomputer and on a
desktop or laptop machine. Documentation and several examples are provided as
Jupyter notebooks using SMURFF's high-level Python API.  The framework has been
successfully used to do large scale runs of compound-activity prediction.

In future version of SMURFF, we will incorporate the use of GASPI and MPI
to be able to run SMURFF efficiently on multiple nodes.

\section*{Acknowledgments}
This work is partly funded by the European projects ExCAPE en EPEEC with
references 671555 and 801051, and by the IT4Innovations infrastructure, which
is supported from the Large Infrastructures for Research, Experimental
Development and Innovations project “IT4Innovations National Supercomputing
Center – LM2015070”.


\end{document}